\documentclass[10pt,twocolumn,letterpaper]{article}

\usepackage{cvpr}
\usepackage{times}
\usepackage{epsfig}
\usepackage{graphicx}
\usepackage{amsmath}
\usepackage{amssymb}

\usepackage[pagebackref=true,breaklinks=true,letterpaper=true,colorlinks,bookmarks=false]{hyperref}

\cvprfinalcopy 

\def\cvprPaperID{9} 

\ifcvprfinal\fi
\begin{document}

\title{DisguiseNet : A Contrastive Approach for Disguised Face Verification in the Wild }

\author{Skand Vishwanath Peri \quad\quad Abhinav Dhall  \\
{\emph{Learning Affect and Semantic Image AnalysIs (LASII) Group,}} \\
{\emph{Indian Institute of Technology Ropar, India}} \\
\small{pvskand@gmail.com, abhinav@iitrpr.ac.in} }

\maketitle

\begin{abstract}
  This paper describes our approach for the \emph{Disguised Faces in the Wild} (DFW) 2018 challenge. The task here is to verify the identity of a person among disguised and impostors images. Given the importance of the task of face verification it is essential to compare methods across a common platform. Our approach is based on VGG-face architecture paired with Contrastive loss based on cosine distance metric. For augmenting the data set, we source more data from the internet. The experiments show the effectiveness of the approach on the DFW data. We show that adding extra data to the DFW dataset with noisy labels also helps in increasing the gen
11
eralization performance of the network. The proposed network achieves 27.13\% absolute increase in accuracy over the DFW baseline.
  
\end{abstract}

\section{Introduction}

Over the years, the research in the area of face recognition has received tremendous amount of attention. Many innovative and novel methods have been put forward for the tasks visual face recognition and verification \cite{Zhao}. Due to its importance, in the past, researchers have concentrated over different problems in face recognition `in the wild'. Here `in the wild' refers to scenarios with varying illumination and pose \cite{facenet}, prediction over varying age of the same person \cite{age}, prediction across different facial expressions \cite{dhall2017individual} and prediction across different modality such as sketches and visual medias etc.\ \cite{klare_mugshot, Ouyang_2016_CVPR}.  

\begin{figure}[!h]
\includegraphics[width=\linewidth]{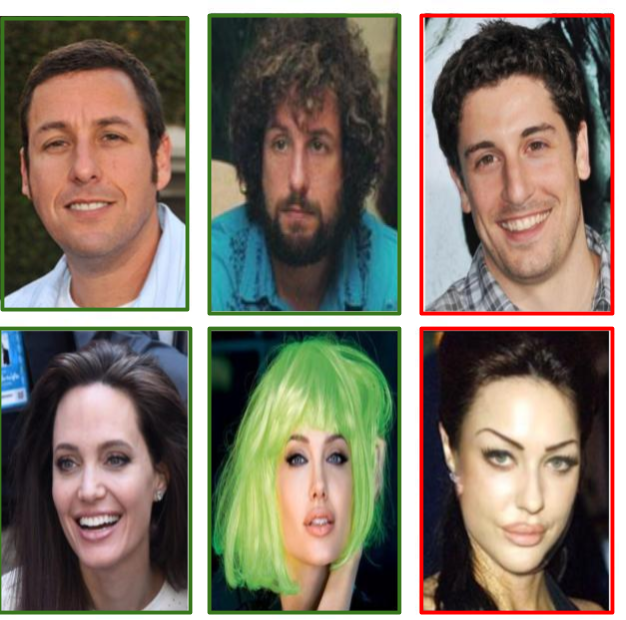}
\caption{
The Figure shows examples of disguised and imposters in the DFW database \cite{dfw}. The first column is a genuine image of a celebrity, the second column is the same person in a disguise and the third columns shows the images of imposters who look like the celebrities in column one. Green bounding box signifies a same identity and red signifies an impostor.}
\vspace{-2mm}
\label{fig:intro_}
\end{figure}

There has been some work on face verification of disguised faces in different imaging modalities like thermal and visual medias \cite{klare_hetero, dhamecha} in which they exploit both the modalities to get the best of both worlds, but the problem of "\textit{disguise}" in a single modality has not been explored in detail. The problem of disguise deals with determining whether the given pair of images are of the same person in disguise or of different persons (one of them being the imposter). 
The Figure \ref{fig:intro_}, clearly show examples of disguised people and imposters.

It was found that successful face recognition systems such as the VGG-Face \cite{Parkhi15} are not efficient, when it comes to the problem of disguise vs impostor recognition \cite{dfw}. VGG-Face achieves 33.76\% Genuine Acceptance Rate (GAR) at 1\% False Acceptance Rate (FAR) and 17.73\% GAR at 0.1\% FAR, which is a clear indication that the usual facial recognition models may not be that helpful to capture the rich representations required for distinguishing the disguised from the imposter images.

Dhamecha et. al.\ \cite{dhamecha}, proposed a two-stage verification process. During the first stage, a patch classifier is computed to decide if a patch is important wrt the task. The hypothesis is that not all facial parts of a disguised person are equally important (particularly occluded parts of a face). The second stage consisted of a patch based face recognition method, in which texture features are extracted from the biometrics-wise important patches. Further, this information is used to verify the personalities with the help of a support vector machine and $\chi^2$ distance metric.

Singh et. al.\ \cite{singh}, used spacial convolutional networks \cite{Pfister15a} to infer the values of fourteen facial key points. Given a disguised image and a gallery of non-disguised images, geometric features are computed based on the angles between the facial key points. The gallery image, which has the least L1 distance based on the geometric feature is assigned as the corresponding non-disguised image of the given disguised image.

Some of the recent face recognition works have also used different loss functions such as the Contrastive loss \cite{hadsell} and the triplet-loss \cite{facenet, Parkhi15} to bring the features representation of the two facial images closer and learn a modality invariant representation.

Disguised faces recognition is an important issue from the perspective of biometrics, as doing this would help surveillance systems recognize imposters trying to steal the identities of other people. The fact that disguised faces increases the \textit{within-class} variation of the faces and imposter faces decreases the \textit{between-class} variation of the faces, makes this task non-trivial. In fact, the imposter images in Figure \ref{fig:intro_} seem to be similar looking to the original identity. Although, when closely observed, we can distinguish between the identities and claim if the person is an imposter or not. From an automated computer vision based method perspective, it is important to extract rich representations of the facial images in-order to distinguish among the identities and verify them correctly.

In recent times, Convolutional Neural Networks (CNN) have been extensively used in computer vision to achieve the state-of-the-art performances \cite{He2015, Simonyan14c} in many classification, object detection tasks and many other vision tasks. CNNs have also provided robust face descriptors \cite{facenet, face_sun}, which have in turn enabled to achieve state-of-art accuracy in face verification and recognition tasks \cite{Wu2015ALC, Richardson_2017_CVPR}.

\begin{figure*}[t]
\centering
\includegraphics[width=145mm]{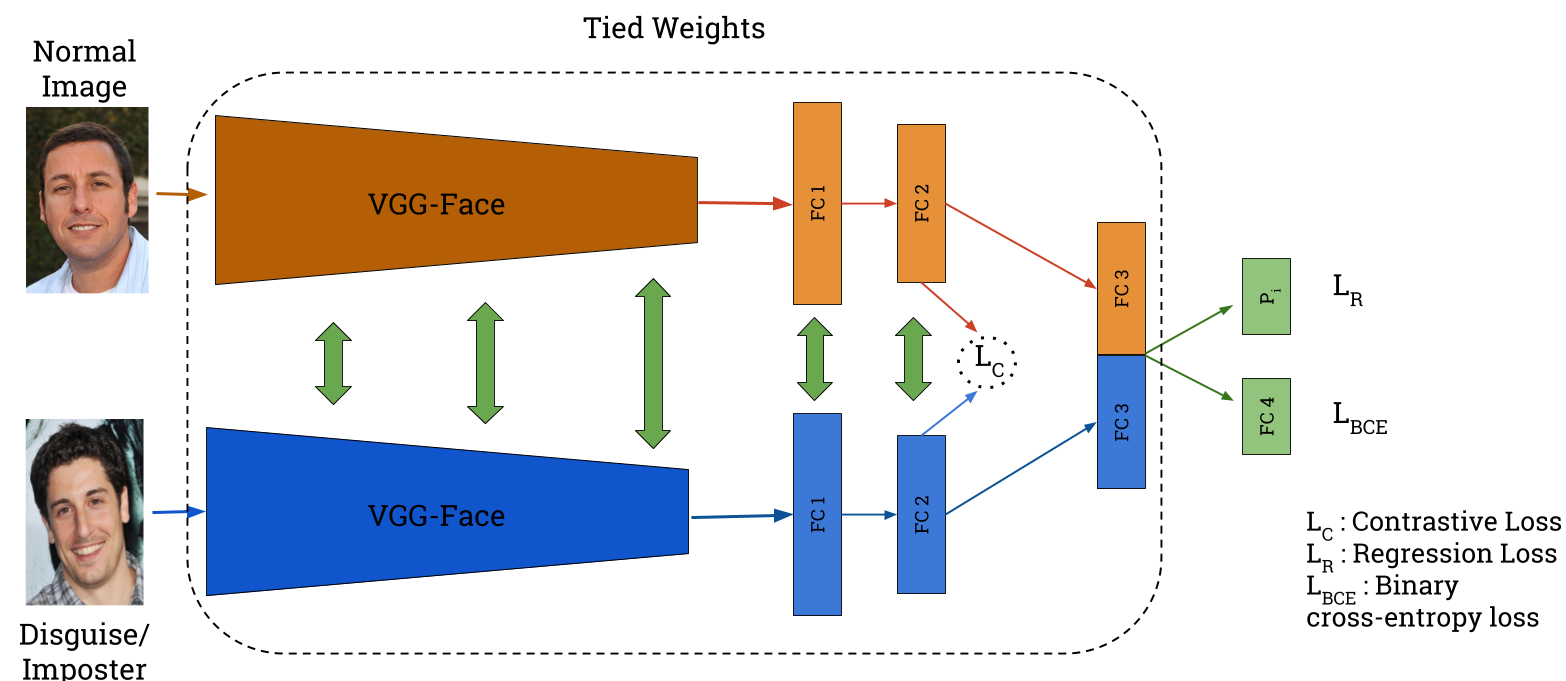}
\caption{
The architecture of the proposed method: Siamese architecture with 16 layered VGG-Face pre-trained weights. A combination of Contrastive Loss ($L_{C}$), Regression Loss ($L_{R}$) and Binary Cross-Entropy Loss ($L_{BCE}$). }
\vspace{-2mm}
\label{fig:arch_}
\end{figure*}

In order to learn rich representations of faces, we use CNNs to extract the features from the images via convolution. To differentiate between the disguised and the imposter images, we explicitly impose Contrastive loss constraint on the feature representation extracted via CNN. This is done so as to bring the representation of the identity and the disguised face closer and to make the representation of the identity and imposter far apart by a margin. Apart from just making the representations far apart, we also regress similarity score of the two images (score = 1, if the images are similar and score = 0, if they are not). The major contributions of the paper are as follows:
\begin{itemize}
\item We propose a CNN model based on VGG-Face \cite{Parkhi15} for the verification of disguised images in the same modality unlike other previous works \cite{dhamecha}, which use CNN for cross-modal verification task of disguised faces.
\item In order to decrease the \textit{intra-class} variation and increase the \textit{inter-class} variation of the feature representation, we use Contrastive loss with cosine similarity measure.
\item To improve the validation accuracy, we also include a regression mean-squared loss apart form the usual classification cross-entropy loss which tells us the similarity of the two images.
\item To increase the performance, we extended the DFW dataset with images from the internet. A total of 1380 images of 325 subjects were downloaded using keyword based search. The new images and their corresponding noisy labels (based on the keyword based search) are used to augment the genuine images in the \emph{Train} set.

\end{itemize}

\section{Disguise Faces in Wild (DFW) Dataset}
We used the Disguise Faces in Wild (DFW) dataset \cite{dfw, dfw2}, which is so far the largest dataset of disguised and imposter images. There have previously been few datasets on makeup disguise in the wild \cite{makeup} and datasets on cross-modality (visual and thermal) disguise faces dataset \cite{iiitd_disguise}. However, to the best of knowledge, there is no dataset, which addresses the issue of disguised faces and imposter faces in the wild. The DFW dataset consists of 1000 identities (400 in training set and 600 in the testing set), with a total of 11155 images. There consists of 3 types of images: genuine images, disguised images and the imposter images. The genuine images are the usual visual image (photograph) of the person, the disguised images is the image of the same identity in disguise and the imposter images are the images consisting of images of other persons who look like the identity. An example of each is shown in Figure \ref{fig:intro_}. We used the coordinates given by the organizers to crop the face out from the image.

\subsection{Weakly Labelled noisy Data from the Internet} The limitation of the DFW dataset is that for every identity, there is only 1 genuine image for training and 1 genuine image for validation. Our experiments suggested that the data with respect to the genuine images (when compared to imposter and disguised images) were less. To overcome the limitation, we downloaded 2-4 more genuine image (totaling the training images to 3-5 for each of the identities in the \emph{Train} set) using the BING image search API \cite{BingSearch} and got a total of 1380 images for 325 celebrities in the DFW \emph{Train} set. We cropped out the faces from the newly added genuine images using the OpenFace library \cite{amos2016openface}. The results of our model with and without the extra data from the internet are mentioned in detail in Table \ref{table2}.

\begin{figure}[!h]
\includegraphics[width=80mm]{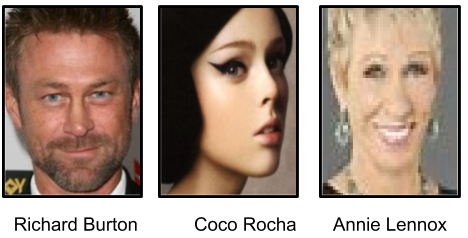}
\caption{
Some of the examples from the extended DFW dataset in which the query returned imposter images.}
\vspace{-2mm}
\label{fig:extra}
\end{figure}

It is interesting to note that the network's performance increased considerably, after the downloaded data was used along with the DFW data. No cleaning of the data was performed and the search engine's results were used to assign the identity to the downloaded images. We noted that in a few cases, images retrieved against the search query generated impostor images too. However, they are used as is in the training without any manual pruning. Some examples of the above mentioned weakly labelled images (which do not belong to the correct identity) are shown in Figure \ref{fig:extra}. We refer to this data as weakly labelled as it is possible that an incorrect identity retrieved during the web search can be due to an attribute of the face in the retrieved image, which makes the image as a good candidate for being the disguised representation of the original identity.

Also it becomes a difficult task for the annotator as well to label the images of the imposter and disguised people. These noisy labels can be generated when the annotator may not be aware about the identity of the subject and other factors such as different ethnicity of the annotator and the subject in the database also add up to the problem of labelling the images.

\subsection{Protocol}
There are three protocols given with the DFW dataset : \emph{Impersonation}, \emph{Obfuscation} and \emph{Overall Performance}. The impersonation protocol consists of image pairs with genuine image and imposter faces, while the obfuscation protocol consists of images of genuine faces with disguised faces. The last protocol consists of all possible pairs (even pair of 2 disguised faces) as a part of the test set.

\section{DisguiseNet\protect\footnote{Code is present at \href{https://github.com/pvskand/DisguiseNet}{https://github.com/pvskand/DisguiseNet}}} 

Let ${ x }_{ i }^{ 1 }$ denote the genuine image of an identity $i$ and ${ x }_{ i }^{ 2 }$ be either the disguised image of identity $i$ or the imposter of identity $i$. We have a Siamese VGG-Face architecture i.e. two sets of 16-layered VGG-Face pre-trained network with tied weights to extract the same features from the input pairs. Note that we need same features as the modality of the images is the same (visual modality). However, the feature representations will be different for genuine image and the disguised/imposter image. We pass the genuine image from the first stream of CNN network and either disguised image or imposter image through the second CNN stream.

Let ${F}_{i}^{m}$ be the feature descriptor of $i^{th}$ identity and $m$=\{\emph{genuine} / \emph{imposter} / \emph{disguise}\}, then 
${ F }_{ i }^{m}=Conv\left( { x }_{ i }^{ m },\Theta  \right) $, where $\Theta$ is the convolutional net parameters. Further, for a Siamese network, the convolutional net parameters: $\Theta$ is the same for all the $m$. Since we use 16 layer deep CNN, the first four convolutional layers are frozen during the fine tuning. The rationale behind this being the observation of no significant gradient flow through the initial layers during the fine tuning process. We have used the Contrastive loss at the second last fully connected layer, which is defined as follows:

\begin{equation}
    L_{C} = \frac { 1 }{ 2B } \sum _{ i=1 }^{ B }{ \left( { y }_{ i }{ d }^{ 2 }+\left( 1-{ y }_{ i } \right) \max { { \left( margin-d,\quad 0 \right)  }^{ 2 } }  \right)  } 
\end{equation}

where $B$ is the batch size, $d$ is a similarity score. In our experiments, cosine similarity distance metric is used for computing $d$. Further, $y_{i}$ is the label of the pair, with $y_{i}=1$ signifying a (genuine, disguise) pair and $y_{i}=0$ signifying a (genuine, imposter) pair. Here, the parameter $margin$ is a scalar value representing the minimum desired distance between a negative (imposter) and positive (disguised) sample.

The second loss, which we include in our network is the verification binary cross-entropy loss. We added this classification task as a regularizer to the Contrastive loss similar to the classification network used as regularizer in \cite{google_net}. 

Learning directly the hard label (0/1) can be a problem in data with such wide intra-class and less inter-class variance. So we use the regression based mean-squared error loss between the predicted score and the actual pair label (0/1). We found that doing this was indirectly acting as a regularizer as well as forcing the representations of same labels to be similar.

\begin{equation}
    L_{R} = { \left\| { y }_{ i }-{ p }_{ i } \right\|  }^{ 2 }
\end{equation}

where $y_{i}$ is the ground truth label and $p_{i}$ is the predicted value after sigmoid activation. We use sigmoid to make sure that the regressed value is between 0 and 1.

Our final loss function is as follows:
\begin{equation}
    L = L_{C} + L_{R} + L_{BCE}
\end{equation}
where $L_{BCE}$, is the binary cross-entropy loss for verification. The contrastive loss is applied on FC2 and hence affecting the weights of all the previous layers of the network and the BCE loss and MSE loss is applied at the end of the network and affects all the layers of the network (except the first four layer as they are freezed).
Also since the positive (genuine, disguise) and negative (genuine, imposter) examples generated from the training dataset is not equal (negative samples are little less than the positive samples as some of the identities don't have imposter images but have only disguised images), so we also use class balancing in the loss i.e weight the negative samples' loss more and weight the positive samples' less. The weights are inversely proportional to number of corresponding samples present in the training data. The architecture is shown in Figure \ref{fig:arch_}.

\begin{table}[t]
  \begin{center}

    \begin{tabular}{|c|c|} 
    \hline
      \textbf{\emph{margin}} & \textbf{Validation Accuracy (\%)}\\
      \hline
      0.1 & 74.59 \\ \hline 
      0.5 & \textbf{79.86} \\ \hline 
      0.6 & 76.07\\
      \hline 
    \end{tabular} 
    \bigskip
    \caption{Parameter analysis of margin. Note that the validation accuracy is on the \emph{Validation} set created by us from the DFW \emph{Train} set.}
    \label{table1}
  \end{center}
\end{table}

\section{Experimental Results \& Ablations}
In order to increase the dataset size for training and also to make our model robust to slight noise and augmentation, we applied data augmentation techniques such as adding Gaussian noise, flipping, random rotation and random translation to the images. This lead to significant increase in the validation accuracy on the test set. We used stochastic gradient descent  optimization in all our experiments with a learning rate of $10^{-3}$. Apart from the DFW \emph{Test} dataset, we also selected random pairs of genuine and disguise/imposter images as our \emph{Validation} set. We show the results of all the ablations on this \emph{Validation} set. We later the compare the results of experiments with the organizer's test set. 

\begin{table}[t]
  \begin{center}
    
    \scalebox{0.95}{
    \begin{tabular}{|l|c|} 
    \hline
      \textbf{Data Setting} & \textbf{Validation Accuracy (\%)}\\
      \hline
      w/o weakly labelled data & 60.43 \\ \hline 
      Weakly labelled data + DFW & \textbf{79.86} \\
      \hline 
    \end{tabular} }
    \bigskip
    \caption{Effect of data augmentation on \emph{validation} accuracy. Note that the validation accuracy is on the \emph{Validation} set created by us from the DFW \emph{Train} set.}
    \label{table2}
  \end{center}
\end{table}

\begin{table}[b]
  \begin{center}
    
    \scalebox{0.92}{
    \begin{tabular}{|l|c|c|c|} 
    \hline
      \textbf{Loss} & \textbf{GAR @0.1\%} & \textbf{GAR@1\%}& \textbf{GAR@10\%}\\
             & \textbf{FAR} & \textbf{FAR}& \textbf{FAR}\\
      \hline
      Baseline \cite{dfw}  & 17.73 & 33.76 & $-$ \\ \hline 
      $L_{C} (W)$ & 19.54 & 50.16 & 92.46\\ \hline 
      $L_{C} + L_{R} (W)$ & 19.71 & 51.53 & 93.86\\ \hline 
      $L_{C} + L_{R}$ & 19.83 & 51.94 & 94.28\\ 
       $+ L_{BCE} (W)$ &  &  & \\\hline 
      $L_{C}$ & 21.03 & 58.32 & 97.86\\ \hline 
      $L_{C} + L_{R}$ & 21.32 & 58.46 & 98.05\\ \hline 
      $L_{C} + L_{R}$ &\textbf{23.25} & \textbf{60.89} & \textbf{98.99}\\
        $+ L_{BCE}$ &  &  & \\
      \hline 
    \end{tabular}}
    \bigskip
    \caption{GAR values at different FAR values on the DFW \emph{Test} set. The baseline method extracted features from VGG-Face model and compared the features with cosine similarity metric. The methods with \emph{(W)} are the methods with the given loss function without the weakly labelled data. }
    \label{table4}
  \end{center}
\end{table}

\subsection{Ablations on Contrastive loss margin}

On changing the value of $margin$, we could get the best validation accuracy at $0.5$. We did not go beyond $margin=1$ as the similarity metric, we are using is cosine similarity, which always returns a value between [0, 1]. The quantitative results are shown in Table \ref{table1}. Note that in the table, we use accuracy metric. This was used for tuning the model parameters on a \emph{Validation} set created by us from the DFW \emph{Training} data.

\subsection{Is Weakly Labelled Data necessary?}
Yes, because it is extremely evident from the below experiment that doing adding the extra dataset with weak labels helps the model to extract robust features in-turn boosts up the performance by a large margin. The results of this ablation are shown in Table \ref{table2} on the \emph{Validation set} created from the DFW \emph{Train set} and the results on DFW Test set are shown in Table \ref{table4}. We can see a significant increase in the accuracy after adding the weakly labelled data in both the tables. On the DFW \emph{Test Set} there is an increase of $~10\%$ for GAR@1\%. All the mentioned experiments are done with $margin=0.5$.

\subsection{Ablation with different loss functions}
We did experiments with Contrastive Loss ($margin=0.5$), Regression loss and binary cross-entropy loss with different combinations of each. It is clear that having regression loss and binary cross-entropy loss act as a regularizer. The results of this are present in Table \ref{table3_1}.

\begin{table}[t]
  \begin{center}
    
    \label{table3_1}
    \begin{tabular}{|l|c|} 
    \hline
      \textbf{Loss} & \textbf{Validation Accuracy (\%)}\\
      \hline
      $L_{C}$ & 76.45 \\ \hline 
      $L_{C} + L_{R}$ & 78.95 \\ \hline 
      $L_{C} + L_{R} + L_{BCE}$ & \textbf{79.86} \\
      \hline 
    \end{tabular} 
    \bigskip
    \caption{Effect of using different combinations of loss functions on the \emph{Validation} set.} \vspace{-2mm}
    
  \end{center}
\end{table}

\begin{figure}[b]
\includegraphics[width=\linewidth]{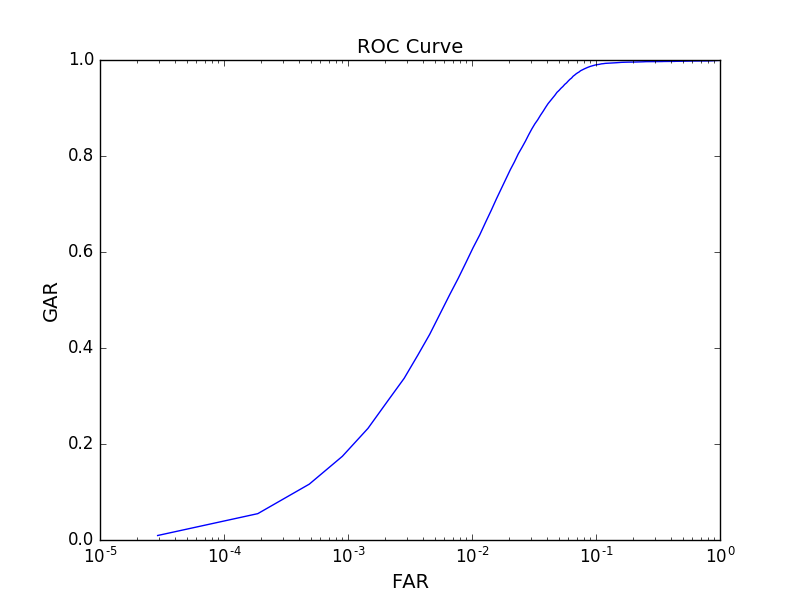}
\caption{
ROC Curve for the proposed model.}
\vspace{-2mm}
\label{fig:roc_disguise}
\end{figure}

\subsection{Evaluation on the Test set}
The results of DisguiseNet on the Test set (provided by the organizers) is shown in Table \ref{table4}. The GAR @ 10\% FAR is close to 99\%  and GAR@1\% FAR is around 61\%. The results of the model with different combinations of loss function is shown and it has turned out that have the Regression and Binary Cross-Entropy Loss along with the Contrastive Loss has helped the model out-perform the other models. The Figure \ref{fig:roc_disguise}, shows the ROC curve for the proposed model.

\begin{figure}[t]
\includegraphics[width=\linewidth]{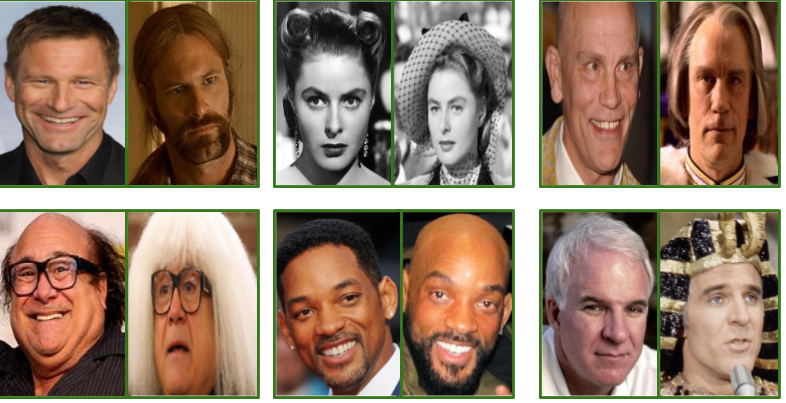}
\caption{
The Figure shows eg.\ of \textbf{True Negative} i.e. both the images in each of the six pairs are of the same identity (the one on the left is a genuine image and the one on the right is a disguised image), but our model says that the two images are that of an imposter images.}
\vspace{-2mm}
\label{fig:mis1}
\end{figure}

\begin{figure}[b]
\includegraphics[width=\linewidth]{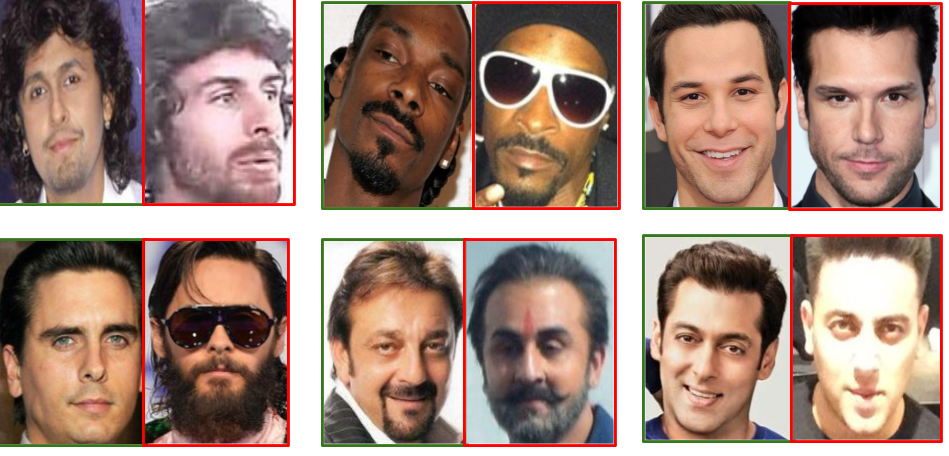}
\caption{
The Figure shows eg.\ of \textbf{False Positives} i.e. both the images in each of the six pairs are of the different identity (the one with green border is the genuine image and the one with the red border is the imposter image), but our model says that the two images are that of the same identity.}
\vspace{-2mm}
\label{fig:mis2}
\end{figure}

Some failure cases of our model for various pairs are shown in Figure \ref{fig:mis1} and Figure \ref{fig:mis2}. The Figure \ref{fig:mis1} shows the cases of \textbf{True Negative} i.e. the pairs shown are actually of the same identity but our model has classified them wrongly. In Figure \ref{fig:mis2}, examples of \textbf{False Positives} are shown where in the images are actually of different people but our networks has classified them as the same identity.

\begin{figure}[t]
\includegraphics[width=\linewidth]{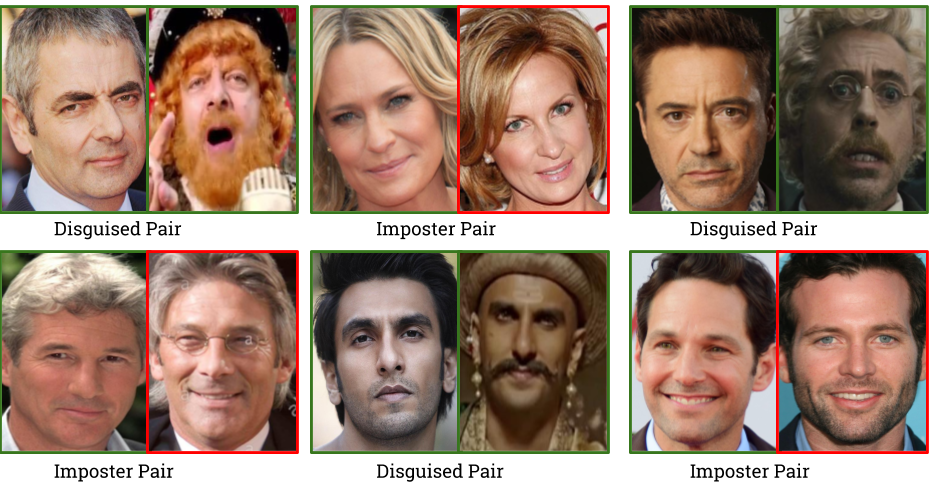}
\caption{
The Figure shows examples of \textbf{False Negatives} and \textbf{True Positives} i.e the both the images either correspond to imposter pair (the pairs with green and red borders) or disguised pair (the pairs with only green border) and our model has predicted the relationship correctly.}
\vspace{-2mm}
\label{fig:true}
\end{figure}

Our success cases are shown in Figure \ref{fig:true}, in which we show both \textbf{False Negatives} and \textbf{True Positives}. The pairs highlighted in green are of the same identity, one being the genuine image and the other being the disguised image whereas the pairs in which one of the images is highlighted in green and the other in red belong to different identities. In all the image pairs shown, our model has correctly classified if they are disguised or imposter.

\section{Conclusion}
In this paper, we proposed a Contrastive loss based approach for verification of disguised faces in the wild (DisguiseNet). The method exploits, the usage of three different loss functions. It is evident from the experiments that the ensemble of three loss functions is beneficial towards the task. Apart from the DFW data, we also add our own data with weak noisy labels in order to enhance the performance of the model for disguise detection. The increase in the data and the use ensemble of loss functions reflects positively on the final results on the DFW \emph{Test} set, giving an absolute increase of 27.13\% over the DFW baseline.

\section{Acknowledgements}
We would like to thank NVIDIA Corporation for donating the TitanXP GPU used for this research.

{\small 
\bibliographystyle{ieee}
\bibliography{disguisenet}

\begin{thebibliography}{10}\itemsep=-1pt

\bibitem{BingSearch}
Bing image search api.
\newblock
  \url{https://azure.microsoft.com/en-us/try/cognitive-services/?api=bing-image-search-api}.
\newblock Accessed: 2018-02-25.

\bibitem{iiitd_disguise}
\emph{IIIT-D} disguise version 1 face database.
\newblock \url{http://iab-rubric.org/resources/facedisguise.html}.

\bibitem{makeup}
Makeup in wild database.
\newblock \url{http://www.antitza.com/makeup-datasets.html}.

\bibitem{amos2016openface}
B.~Amos, B.~Ludwiczuk, and M.~Satyanarayanan.
\newblock Openface: A general-purpose face recognition library with mobile
  applications.
\newblock Technical report, CMU-CS-16-118, CMU School of Computer Science,
  2016.

\bibitem{age}
D.~Deb, L.~Best-Rowden, and A.~K. Jain.
\newblock Face recognition performance under aging.
\newblock In {\em {Proceedings of the IEEE Conference on Computer Vision and
  Pattern Recognition Workshops (CVPRW)}}, pages 548--556, July 2017.

\bibitem{dhall2017individual}
A.~Dhall, R.~Goecke, S.~Ghosh, J.~Joshi, J.~Hoey, and T.~Gedeon.
\newblock From individual to group-level emotion recognition: Emotiw 5.0.
\newblock In {\em Proceedings of the 19th ACM International Conference on
  Multimodal Interaction}, pages 524--528. ACM, 2017.

\bibitem{dhamecha}
T.~Dhamecha, A.~Nigam, R.~Singh, and M.~Vatsa.
\newblock Disguise detection and face recognition in visible and thermal
  spectrums.
\newblock In {\em {Proceedings of the International Conference on Biometrics
  (ICB)}}, pages 1--8, 06 2013.

\bibitem{dfw2}
T.~I. Dhamecha, R.~Singh, M.~Vatsa, and A.~Kumar.
\newblock Recognizing disguised faces: Human and machine evaluation.
\newblock {\em PLOS ONE}, 9(7):1--16, 07 2014.

\bibitem{hadsell}
R.~Hadsell, S.~Chopra, and Y.~LeCun.
\newblock Dimensionality reduction by learning an invariant mapping.
\newblock {\em {Proceedings of the IEEE Conference on Computer Vision and
  Pattern Recognition (CVPR)}}, 2:1735--1742, 2006.

\bibitem{He2015}
K.~He, X.~Zhang, S.~Ren, and J.~Sun.
\newblock Deep residual learning for image recognition.
\newblock {\em arXiv preprint arXiv:1512.03385}, 2015.

\bibitem{klare_hetero}
B.~Klare and A.~K.~Jain.
\newblock Heterogeneous face recognition: Matching nir to visible light images.
\newblock pages 1513--1516, 08 2010.

\bibitem{klare_mugshot}
B.~Klare, Z.~Li, and A.~K~Jain.
\newblock Matching forensic sketches to mug shot photos.
\newblock In {\em {IEEE Transactions on Pattern Analysis and Machine
  Intelligence}}, volume~33, pages 639--46, 09 2010.

\bibitem{dfw}
V.~Kushwaha, M.~Singh, R.~Singh, M.~Vatsa, N.~Ratha, and R.~Chellappa.
\newblock {Disguised Faces in the Wild}.
\newblock Technical report, IIIT Delhi, March 2018.

\bibitem{Ouyang_2016_CVPR}
S.~Ouyang, T.~M. Hospedales, Y.-Z. Song, and X.~Li.
\newblock Forgetmenot: Memory-aware forensic facial sketch matching.
\newblock In {\em {Proceedings of the IEEE Conference on Computer Vision and
  Pattern Recognition (CVPR)}}, 2016.

\bibitem{Parkhi15}
O.~M. Parkhi, A.~Vedaldi, and A.~Zisserman.
\newblock Deep face recognition.
\newblock In {\em British Machine Vision Conference}, 2015.

\bibitem{Pfister15a}
T.~Pfister, J.~Charles, and A.~Zisserman.
\newblock Flowing convnets for human pose estimation in videos.
\newblock In {\em {Proceedings of the IEEE International Conference on Computer
  Vision (ICCV)}}, 2015.

\bibitem{Richardson_2017_CVPR}
E.~Richardson, M.~Sela, R.~Or-El, and R.~Kimmel.
\newblock Learning detailed face reconstruction from a single image.
\newblock In {\em {Proceedings of the IEEE International Conference on Computer
  Vision and Pattern Recognition (CVPR)}}, July 2017.

\bibitem{facenet}
F.~Schroff, D.~Kalenichenko, and J.~Philbin.
\newblock Facenet: A unified embedding for face recognition and clustering.
\newblock In {\em {Proceedings of the IEEE Conference on Computer Vision and
  Pattern Recognition (CVPR)}}, pages 815--823, 2015.

\bibitem{Simonyan14c}
K.~Simonyan and A.~Zisserman.
\newblock Very deep convolutional networks for large-scale image recognition.
\newblock {\em CoRR}, abs/1409.1556, 2014.

\bibitem{singh}
A.~Singh, D.~Patil, G.~Reddy, and S.N.Omkar.
\newblock Disguised face identification with facial keypoints using spatial
  fusion convolutional network.
\newblock In {\em {Proceedings of the IEEE Conference on Computer Vision and
  Pattern Recognition Workshop}}.

\bibitem{face_sun}
Y.~Sun, X.~Wang, and X.~Tang.
\newblock Deep learning face representation from predicting 10,000 classes.
\newblock In {\em Proceedings of the IEEE Conference on Computer Vision and
  Pattern Recognition}, 2014.

\bibitem{google_net}
C.~Szegedy, W.~Liu, Y.~Jia, P.~Sermanet, S.~Reed, D.~Anguelov, D.~Erhan,
  V.~Vanhoucke, and A.~Rabinovich.
\newblock Going deeper with convolutions.
\newblock In {\em Computer Vision and Pattern Recognition (CVPR)}, 2015.

\bibitem{Wu2015ALC}
X.~Wu, R.~He, and Z.~Sun.
\newblock A lightened cnn for deep face representation.
\newblock {\em CoRR}, abs/1511.02683, 2015.

\bibitem{Zhao}
W.~Zhao, R.~Chellappa, P.~J. Phillips, and A.~Rosenfeld.
\newblock Face recognition: A literature survey.
\newblock {\em ACM Comput. Surv.}, 35(4):399--458, Dec. 2003.

\end{thebibliography}
}

\end{document}